\documentclass[conference,letter,twocolumn]{IEEEtran}

\usepackage{graphicx}  

\begin{document}

\newcommand{\p}{\textbf{p}}
\newcommand{\q}{\textbf{q}}
\newcommand{\D}{\textbf{D}}
\newcommand{\A}{\textbf{A}}
\newcommand{\bb}{\textbf{b}}

\title{\Huge Planning Random path distributions for ambush games in unstructured environments}

\author{ \authorblockN{\Large Emmanuel Boidot}
\authorblockA{\large Georgia Institute of Technology\\
Atlanta, Georgia, USA\\
emmanuel.boidot@gatech.edu}  \and
\authorblockN{\Large Eric Feron}
\authorblockA{\large Georgia Institute of Technology\\
Atlanta, Georgia, USA\\
feron@gatech.edu} }


\maketitle

\begin{abstract}
Operating vehicles in adversarial environments require non-conventional planning techniques. A two-player, zero-sum non-cooperative game is introduced, and solved via a linear program. An extension is proposed to construct networks displaying good representations of the environment characteristics, while offering acceptable results for the technique used. Sensitivity of the solution to the LP solver algorithm is identified. The planner's performances are finally assessed by comparison with those of conventional planners. Results are used to formulate secondary objectives to the problem. 
\end{abstract}

\vspace{0.2cm} {\bfseries{\textit{\small Keywords}}}: \ {\it{Path planning, linear optimization, ambush games, probability distributions}}

%

\section{Introduction}
Most of us travel repeatedly between the same departure and arrival places. For most applications, conventional path planning are sufficient to determine how to get from one point to another. In a hostile environment, such journeys can be threatened, for instance by vehicle ambushes. These ambushes might have different purposes, for instance stealing the content of a transport or attacking security officers, but in all cases the assessment is the same: following the same path repeatedly from one occurrence to the next is not a good idea, because it helps adversaries make their attacks most efficient.

While ambush games have been classical examples of game theory problems for a long time now, there have been only a few attempts to solve ambush games with practical applications in mind. The idea of such an application of operations research is presented in \cite{Ergun}.
Joseph and Feron \cite{Farmey-thesis,feron:0001} model this problem as a zero-sum non-cooperative two players game. Considering a roadmap, their goal is to minimize the expected penalty of getting ambushed over the probability vector representative of the stochastic behavior of a convoy at each intersection on the road, the output of this model being the corresponding optimal vector.
Salani, Duyckaerts and Schwartz \cite{SalaDuykSwart2010} adapted the model in \cite{Farmey-thesis} for convoys that must stop at multiple locations, for instance in the case of money distribution vehicles. The works above do not assume any information about the opponent position. This a relatively good assumption for ground vehicles in hostile environment with possible attacks from any direction. 
Other approaches suppose more information about the opponent. For example, Karaman and Frazzoli \cite{Kara-Frazzo} use a hierarchical (Stackelberg) game to model the problem, where the convoy has information about the initial position of the opponent(s). The model is used to construct a dynamic pursuit-evasion game where the opponent(s) can change their strategy during the game.


The end goal of this project is to develop both strategic and dynamic ambush avoidance algorithms, i.e. global route decision and real time maneuvers, that would increase humanitarian convoy safety in hostile environments.  
In this paper, the problem of reducing the risk incurred by a convoy along its entire path is approached by using the model developed in \cite{Farmey-thesis}. Only single-stage single-ambush games are addressed, where both players decide of their strategy before the beginning of the games. 

Section II presents the model that supports this research. Section III extends the model to unstructured environments in order to make it more practical and realistic. In Section IV the set of optimal solutions to the formulated linear problem problem is studied, leading to the description of secondary objectives. 

\section{Approach}

	\subsection{Game Description}
The problem of interest is to plan a path for a convoy that needs to journey from point A to point B in a given environment. It is modeled as a two players non-cooperative zero-sum game where Player 1 runs the convoy that Player 2 tries to ambush. 
The environment is described by a network $(N,E)$ and a risk map. Ambushes take place at intersection (nodes) of the network. Each vertex $n_i$ is associated with a real value $\alpha_i$ that represents the outcome for Player 2 if he sets an ambush on vertex $n_i$ and Player 1 path goes through it. The set \textbf{$\alpha$} is the risk map, as it measures the possible losses for player 1.
		
A possible strategy for Player 1 is represented by a probability vector $\p$ that contains the probability $p_{ij}$ that the convoy uses edge $e_{ij}$ between nodes $n_i$ and $n_j$. Similarly, a strategy for Player 2 is represented by a probability vector $\q$ that contains the probability $q_j$ that he sets an ambush at node $n_j$.		
The goal of the game is, given the network and the risk map,to find the optimal strategy $\p^{\ast}$ for Player 1, assuming that Player 2 follows its optimal strategy $\q^{\ast}$.
It is likely to be a mixed strategy: Player 1 and Player 2 strategies are not deterministic. Instead they are probabilistic and the path followed by the convoy as well as the location of the ambush may be different for a same strategy $\p$ or $\q$.

	\subsection{Mathematical Formulation}
Let $(N,E)$ be a network with $n$ edges and $m$ nodes. Let $\p$ (resp. $\q$) be the probability vector representative of Player 1's (resp. Player 2's) mixed strategy.

Assume that the two players strategy are independent. At each node $n_j$, the probability that Player 1 gets ambushed is equal to the probability that Player 1's path goes through $n_j$ times the probability that Player 2 sets an ambush at this node. The gain for Player 2 at this node being $\alpha_i$, the expected outcome of the game relative to this node is: $\sum\limits_{i | (i,j)\in E} p_{ij} q_j \alpha_j$. Therefore the global outcome of the game is
\begin{equation}
V = \sum\limits_{j \in N} \sum\limits_{i | (i,j)\in E} p_{ij} q_j \alpha_j \\
		= \q^t \D \p. 
\end{equation}
with $D_{jk} = \alpha_j$ if the $k^{th}$ line of p represents the probability that Player 1 uses an edge $e_{ij}$ directed towards $n_j$, and $D_{jk} = 0$ otherwise.

As to minimize the risk encountered by the convoy, the objective of the approach is to find the strategy for Player 1 that minimizes the maximal possible outcome for Player 2:

\begin{equation}
\p^{\ast} = \arg\min\limits_{\p} \max\limits_{\q} \q^t \D \p. 
\end{equation}

Providing the fact that $q_j \leq 1\ \ \forall j$, Player 2 can always maximize V by choosing the node $n_j$ for which the probability of Player 1 passing through that node weighted by the value $\alpha_j$ is maximal. Therefore Player 1's optimal solution is to minimize this product across all nodes:

\begin{equation}
\p^{\ast} = \arg\min\limits_{\p} \left( \max\limits_{j \in N} \sum\limits_{i | (i,j) \in E} p_{ij} \alpha_j \right). 
\end{equation}

The other constraints of this problem illustrate the flow conservation inside the network. The probability of the convoy arriving at node $n_j$ is equal to the probability of the convoy leaving this node. Probabilities of the convoy being at origin and destination nodes are equal to 1.

\begin{equation}
\left \{
\begin{array}{c @{=} lr}
    \sum\limits_{i | (i,j) \in E} p_{ij} & \sum\limits_{k | (j,k) \in E} p_{jk},& \forall j \in N \setminus \{n_0,n_d\}\\
    \sum\limits_{j | (n_0,j) \in E} p_{n_0j} & \ \ \ 1&\\
    \sum\limits_{j | (j,n_d) \in E} p_{jn_d} & \ \ \ 1&\\
\end{array}
\right.
\end{equation}

This problem is solved as a linear optimization problem by introducing a variable $z$ constrained as follow.
\begin{equation}
\begin{array}{c @{\geq} lr}
z\ & \sum\limits_{i | (i,j)\in E} p_{ij} \alpha_j & \forall j \in N
\end{array}
\end{equation}  

Rewriting Equations (3), (4) and (5) the Linear Problem can be written under the following matrix form, with A and b representing the flow conservation constraints.
\begin{equation}
\begin{array}{|rl|}
\hline
\multicolumn{2}{|c|}{$minimize$\ z}\\
$subject to$\ \ \ \ \D\p - $\textbf{1}$z & \leq 0\\
\A\p & = \bb \\
\p & \geq\ $\textbf{0}$ \\ \hline
\end{array}
\end{equation}

	\subsection{Example}	
\begin{figure}
\begin{center}
	\includegraphics[scale=0.5]{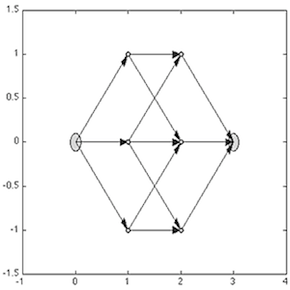}
	\vspace*{-1em}
	\caption{A trivial network example.}
	\label{simple_network}
\end{center}
\vspace*{-1em}
\end{figure}

A example of very simple network (8 nodes, 13 edges) is displayed in Figure \ref{simple_network}. Edges are directed so that the vehicle can only move in the directions shown by the arrows. The cost $\alpha$ of an ambush is assumed to be equal to one for each internal node and zero on departure and arrival nodes.

\begin{figure}
\begin{center}
	\includegraphics[scale=0.5]{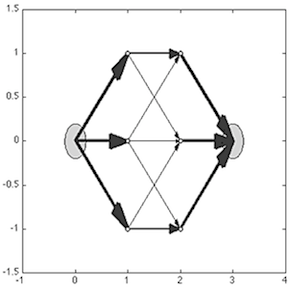}
	\vspace*{-1em}
	\caption{Optimal strategy (interior-point solution) for the network displayed in Figure \ref{simple_network}. The wider the arrow is, the higher the probability that Player use this edge is.}
	\label{simple_network_sol}
\end{center}
\vspace*{-1em}
\end{figure}

The result of the optimization technique presented above is displayed in Figure \ref{simple_network_sol}. The probability of each edge being used is computed as to minimize expected losses. This probability is represented as the width of each edge. Using a symmetric, alpha-uniform network, an interesting result is obtained: the probability of the vehicle passing by is well spread across the network and preserve the initial symmetry. This property is very interesting regarding the goal of this technique,  which is to avoid ambushes. It seems to indicate that optimal solutions to the problem are also the most deceptive ones because most paths are as likely. Note that the linear solver used for this optimization was the interior-point algorithm.

\section{Network construction}

	\subsection{Method description}
Though this approach is thoroughly described in \cite{Farmey-thesis}, it assumes the prior existence of a network to optimize on. Furthermore, no description is provided of the sensitivity of the solution towards the alpha distribution ("risk map"). An attempt to address these issues is presented in this part.


Assuming the existence of a network limits the number of possible routes for the vehicle. Therefore, it decreases the advantage of this approach that is to increase variability to the convoy's trajectory. Moreover, most ambush situations nowadays take place in warfare zones where the vehicles concerned are more likely to be off-road or aerial vehicles. Hence it is expected that adapting this approach as to create a graph representation of the environment would improve its efficiency. This representation would have to be adapted to the vehicle physical model and environment characteristics.	
	
Several methods were tried in order to create a network that would be as representative of the environment as possible while allowing reasonably fast computation. Differences were made on the sampling method and on the connectivity between nodes as displayed in Table \ref{method_table}. Each link between two nodes requires two oriented edges (one per direction) for its representation. While randomly sampled nodes connected through a Delaunay triangulation result in a relatively small and computationally efficient representation of the environment, Method 1 might not be representative enough of the details of the environment. The best representation of the environment is obtained through Method 2, but it requires the creation of 16 directed edges per node (8 neighboring nodes, 2 directions per edge). This method is computationally intense, therefore it might be preferable to choose a less precise technique that would allow better precision on the environment description. By intensively reducing the number of edges inside the network through simplified connectivity, Method 3 appears as a good alternative for this task.

\begin{table}
\begin{center}
$\begin{array}{|c|c|c|}
\hline
$Method \#$ & $Sampling$ & $Connectivity$ \\
\hline
$1$ & $Random$ & $Delaunay triangulation$\\
$2$ & $Uniform$ & $8 connected grid$\\
$3$ & $Uniform$ & $Delaunay triangulation$\\
%
%
\hline
\end{array}$
	\caption{Different network construction methods.}
	\label{method_table}
\end{center}
\vspace*{-3.5em}
\end{table}

	\subsection{Examples}

\begin{figure}
\begin{center}
$\begin{array}{cc}
\includegraphics[scale=0.5]{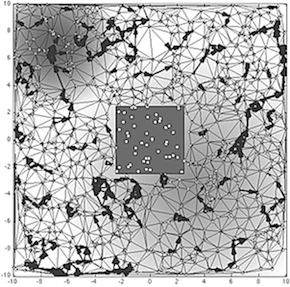} & \includegraphics[scale=0.5]{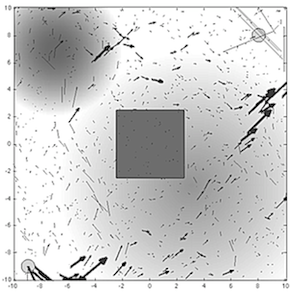}\\
\multicolumn{2}{c}{${\small a. Method 1: Random sampling + Delaunay triangulation}$}\\
\includegraphics[scale=0.5]{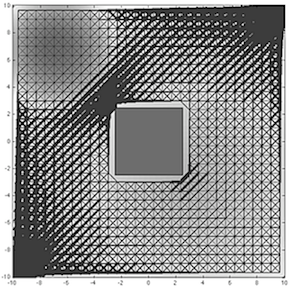} & \includegraphics[scale=0.5]{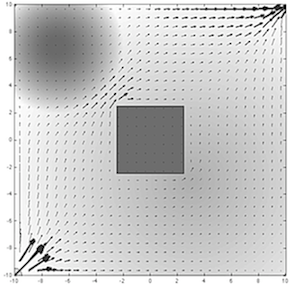}\\
\multicolumn{2}{c}{${\small b. Method 2: Uniform sampling + 8-connected grid}$}\\
\includegraphics[scale=0.5]{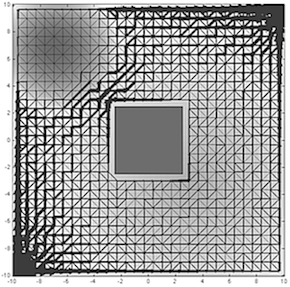} & \includegraphics[scale=0.5]{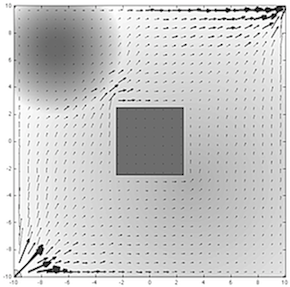}\\
\multicolumn{2}{c}{${\small c. Method 3: Uniform sampling + Delaunay triangulation}$}\\
\end{array}$
	\caption{Optimization results for different network construction methods. The background grayscale represents the casualty forecast \textit{alpha} if the convoy gets ambushed. The gray square in the middle is an obstacle. The left image represents the actual probability for every edge of the network, while the right image represents the mean direction from every node of the network.}
	\label{sampling_results}
\end{center}
\vspace*{-2em}
\end{figure}
	
Figure \ref{sampling_results} shows different optimal solutions obtained on the same environment with networks constructed using the methods described above. It is important to note that a good network leads to results that display features relative to the \textit{alpha} distribution and to the topology of the environment. 
Whereas several very efficient conventional path planning algorithm are based on random sampling of the environment, Figure \ref{sampling_results}.a  shows that the paths generated by Method 1 are rather erratic and do not seem to avoid dangerous area of the environment, which lead to the conclusion that random sampling is not adapted to this specific type of planning. 
As expected, the best results are obtained with Method 2 because it ends up creating a network where all possible paths from the departure node to the arrival node are included. The optimization being realized over all possible paths, the solution obtained is the best given the level of discretization. Therefore it can be assumed that any good representation of the environment should end up with solutions relatively close to solutions obtained in Figure \ref{sampling_results}.b, ie avoiding the areas where \textit{alpha} is high while being well spread across the map. The mean direction figure (right column of Figure \ref{sampling_results}.b) highlights the fact that there are nodes where the probability of going to a specific node is so high that it makes the transition almost deterministic. This is due to the fact that the algorithm leads the convoy to nodes where the hypothetic loss is lower. 
Figure \ref{sampling_results}.c illustrates the good result obtained with the lower connectivity of Method 3. While the probability map might seem slightly different from Figure \ref{sampling_results}.b, the mean direction is almost the same, meaning that at each node the transition probability will be very close. These examples illustrate the good behavior of networks obtained through Method 3 and justify the decision to go on with this method for further studies.

As shown in Figures \ref{sampling_results}.b and \ref{sampling_results}.c, the method returns good results in the sense that many edges all around the environment present similar probability, which means that they are as likely to be used. Our goal being to build a planner that returns less straightforward paths for the opponent to guess, this method succeeds in constructing very deceptive "paths". Not only can the path generated from throwing the dices over these probabilities be almost any path from A to B, it will also vary a lot form one day to another as the dices throw will be reiterated.

Note that current integration of three dimensional topological data leads to the idea that a good alternative to the technique employed here would be to construct a network with a sampling biased towards to elevation gradient and a Delaunay triangulation connectivity. While this is not yet the main subject of interest in this research, this technique could be implemented in the future as an enhancement of the method currently in use. 
	

\section{Evaluation}

	\subsection{Sensitivity to the LP solver}

It has been shown in the analysis of the solution displayed in Figure \ref{simple_network_sol} that a very interesting feature of the method presented in this paper is the high deceptiveness of the optimal solution returned by the planner. This is due to the fact that most edges have a non-zero probability of being used, meaning that almost all possible paths from A to B have a non-zero probability of being used.
However, it was noted that these results were obtained using the \textit{Interior-Point} algorithm to solve the linear optimization problem. 

\begin{figure}
\begin{center}
	\includegraphics[scale=0.5]{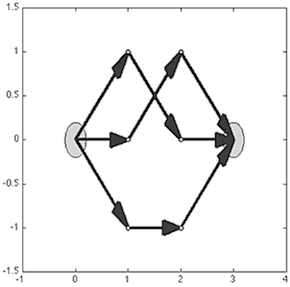}
	\caption{Optimal strategy (simplex solution) for the network displayed in Figure \ref{simple_network}.}
	\label{simple_network_simplex}
\end{center}
\vspace*{-1em}
\end{figure}

\begin{figure}
\begin{center}
$\begin{array}{cc}
\includegraphics[scale=0.5]{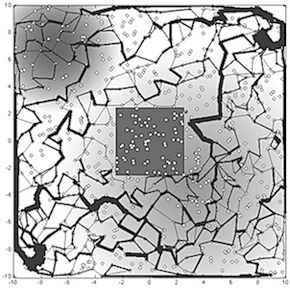} & \includegraphics[scale=0.5]{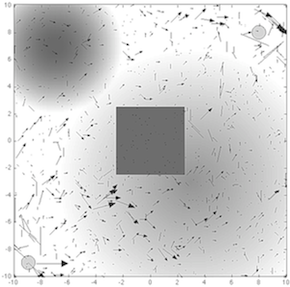}\\
\multicolumn{2}{c}{${\small a. Method 1: Random sampling + Delaunay triangulation}$}\\
\includegraphics[scale=0.5]{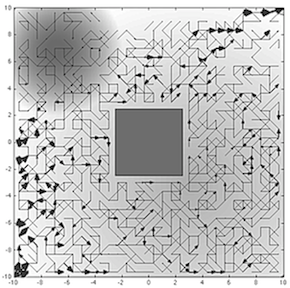} & \includegraphics[scale=0.5]{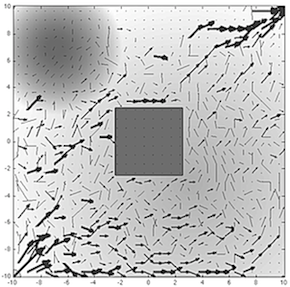}\\
\multicolumn{2}{c}{${\small b. Method 2: Uniform sampling + 8-connected grid}$}\\
\includegraphics[scale=0.5]{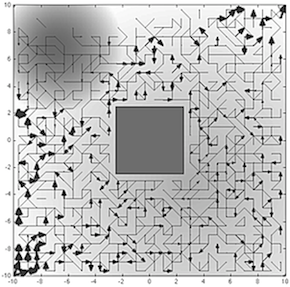} & \includegraphics[scale=0.5]{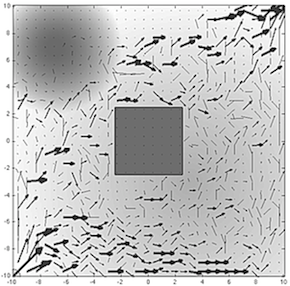}\\
\multicolumn{2}{c}{${\small c. Method 3: Uniform sampling + Delaunay triangulation}$}\\
\end{array}$
	\caption{Optimization results for different network construction methods (simplex algorithm). The background grayscale represents the casualty forecast \textit{alpha} if the convoy gets ambushed. The gray square in the middle is an obstacle. The left image represents the actual probability for every edge of the network, while the right image represents the mean direction from every node of the network.}
	\label{simplex_results}
\end{center}
\vspace*{-1em}
\end{figure}

In order to examine the sensitivity of the planner to other linear optimization algorithm, a similar method is applied to the example in Figure \ref{simple_network} using the \textit{Simplex} algorithm. The result is displayed in Figure \ref{simple_network_simplex}. Instead of spreading across the entire network, thus allowing 7 different paths from A to B, the optimal solution here returned only use 70\% of the network, allowing only 3 different paths from A to B that are completely determined by the first dice throw. This feature is also present when applying the \textit{simplex} algorithm to the networks of Figure \ref{sampling_results}. As shown in Figure \ref{simplex_results}, the portion of network used by the optimal solution is much smaller than in the previous case. While the possible paths from A to B seem much more erratic, the ambush avoidance behavior of the solution is still observable in the mean direction figures. At each node, among all possible following nodes the convoy is most likely to go towards the ones where the possible loss, in case of an ambush, is the lowest.
	
This difference in results between the planner using the \textit{Interior-Point} algorithm and the one using the \textit{Simplex} algorithm is explained by the fact that the \textit{Interior-Point} algorithm is known to lead to solutions maximizing the entropy among the set of optimal solutions. This fact is illustrated in Table \ref{entropy_table}: the \textit{Simplex} algorithm always returns lower entropy solutions.

\begin{table}
\begin{center}
$\begin{array}{|c|c|c|}
\hline
$Network$ & $Interior-Point$ & $Simplex$\\ \hline
1^{st}$ example$ & 4.10 & 3.30\\
$Method 1$ & 934.58 & 117.05\\
$Method 2$ & 331.54 & 244.54\\
$Method 3$ & 404.21 & 239.96\\
\hline
\end{array}$
	\caption{Entropy for different networks}
	\label{entropy_table}
\end{center}
\vspace*{-3.5em}
\end{table}
	
These results lead to two conclusions. The first one is that, given the difference between the solutions obtained through the different LP solver, the set of solutions is reasonably wide. Therefore it is possible to add constraints or secondary objectives to the problem, such as entropy maximization. The second conclusion is that entropy is indeed a very good criteria to evaluate the deceptiveness of the solutions: the more spread the solution is across the network and the higher the entropy is. 

In previous research of Joseph and Feron, the sensitivity of the method to the LP solver used had not been identified, furthermore no evaluation of the solutions quality was realized. Here the entropy of the probability distribution $\p$ was identified as a possible measure for this quality, which justifies the choice of the \textit{Interior-Point} algorithm as the LP solver of the planner.

	\subsection{Performances}
The performances of this technique are evaluated by comparing it to more conventional deterministic planners. The first planner returns the shortest path from A to B on the network. The second planner returns the trivial safer path from A to B on the network, i.e. the one that goes through the nodes with low \textit{alpha}.

The different properties being evaluated over these planners are:
\begin{itemize}
\item Expected energy cost / length E.
\item Probability of ambush for 1 iteration of the game (compute q for each node once solution has been found).
\item Probability of ambush for the $n^{th}$ iteration of the game, $n$ large, when Player has had time to learn the deterministic. strategies.
\item Expected game outcome $V_{\infty}$ for the $n^{th}$ iteration of the game, $n$ large.
\end{itemize}
	
In order to compute the outcome of the game, it is assumed that Player 2 knows Player 1's strategy but cannot guess the results of the dice throw for each game. This is justified by the fact 	that for the technique presented here, Player 1's strategy consists in choosing the optimal strategy assuming that Player 2 does the same. Therefore Player 2 is able guess this strategy. However he is not able to guess the exact path the convoy follows for a specific iteration of the game because the strategy is not deterministic. This makes the convoy safe even if the global strategy is common knowledge. The two other planners being deterministic, it is assumed that Player 2 can determine the complete path after a few iterations of the game, meaning that he will set the ambush as to maximize $V$. For the first iteration Player 2 assumes that Player 1's strategy is the one described in Section II but after a large number of iterations he knows the path taken by the convoy and place his ambush in consequence.

Given the strategy $p$ for Player 1, $q$ is computed as the argument that maximizes the outcome $V$ of the game as defined in Equation (1).
\begin{equation}
\q^{\ast} = \arg\max\limits_{\q} \p^t \D^t \q. 
\end{equation}
For a given iteration of the game, the probability of being ambushed is computed as the joint probability of the convoy going through a node while Player 2 is at this node.
\begin{equation}
P_{ambushed} = \sum\limits_{j \in N} \sum\limits_{i | (i,j)\in E} p_{ij} q_j
\end{equation}
	
\begin{table}
\begin{center}
$\begin{array}{|c|c|c|c|c|} 
\hline
$Planner$ 					& 	$E$ 	& P_1 			& P_{\infty} 	& V_{\infty}\\ 
\hline
$Stochastic - Method 3$		&	89.57	& 0.33 			& 0.33 			& 1.09\\
$Energy efficient$ 			& 	31.24	& 0.33 			& 1 			& 5.99\\
$Risk avoidance$ 			& 	40.87	& 0.33 			& 1 			& 6.43\\
\hline
\end{array}$
	\caption{Comparison to traditional deterministic planners.}
	\label{tab_results}
\end{center}
\vspace*{-3.5em}
\end{table}

Results presented in Table \ref{tab_results} show the performances of the different planners regarding chosen criteria. The technique described in II appears to be energetically inefficient. The expected length of a path issued by this planner is three times the minimal length from A to B. The risk avoidance planner offers decent energy efficiency, being only 30\% less efficient than the cost efficient planner. However this feature is completely dependent on the nature of the environment: if the path of low risk nodes was to wind across the environment, this planner could end up being very inefficient. But the most interesting criteria to compare these planners regard the ambush probability and resulting casualties (outcome of the game). A great feature of the stochastic planner is that even if the opponent ends up choosing the best strategy possible, he is not able to ambush the convoy with a probability of 1. In fact, the probability of being ambushed at a given iteration cannot exceed the probability of being ambushed when Player 2 chooses its optimal strategy. As the paths returned by the planner are random there is no possibility for the opponent to learn more than the probability distribution of Player 1's optimal strategy. On the other hand, when facing deterministic planners Player 2 can quickly learn the path used by the convoy and place its ambush on this path. Therefore the probability of the convoy being ambushed converges to 1 with the number of iterations of the game, meaning that it ends up being ambushed every time it goes from A to B. Moreover, the expected loss to Player 1 when his convoy is ambushed is much more important with the deterministic planners. 
	
\begin{figure}
\begin{center}
	\includegraphics[scale=0.5]{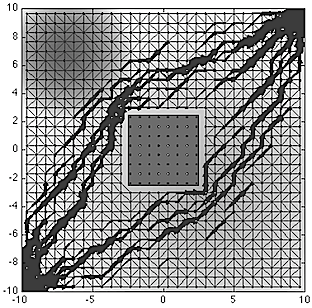}
	\caption{Secondary objective improvement: Energy efficiency.}
	\label{improve_energy}
\end{center}
\vspace*{-1em}
\end{figure}

	
While the energy efficiency of the method is not satisfying, it is possible to modify the optimization in order to make it much more efficient. For example by minimizing $0.99z + 0.01\sum\limits_{j \in N} \sum\limits_{i | (i,j)\in E} p_{ij}l_{ij}$ instead of minimizing $z$, the expected length of the path decreases from 89.57 to 33.14, close to the optimal value. This improvement occurs at the expense of the safety constraints, as shown in Figure \ref{improve_energy}. Probabilities have increased in for edges in the SE area where \textit{alpha} is high because shorter path goes through this area.

\begin{figure}
\begin{center}
	\includegraphics[scale=0.5]{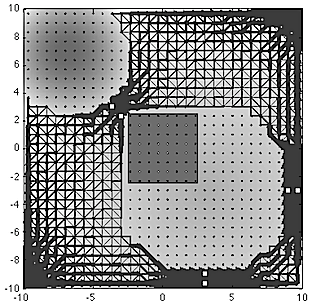}
	\caption{Secondary objective improvement: Safety. Little white squares represent the optimal strategy for Player 2.}
	\label{improve_safety}
\end{center}
\vspace*{-1em}
\end{figure}

Another improvement possibility is to forbid areas of the environment where the cost of being ambushed would be to high. By discarding nodes $n_i$ for which $\alpha_i > \alpha_{threshold}$, a reduced network is obtained. While this might be an interesting improvement to the technique, a major drawback is that, if the threshold is chosen too low, all paths merge into a few bottlenecks, as shown in Figure \ref{improve_safety}. The danger of these bottlenecks is that they ease the opponent task, which is to choose nodes where the convoy is more likely to go through. Because of this difficulty to choose a good value for \textit{alpha}, it might not be recommended to use this feature. 

\vspace*{1em}	
Using deterministic planners becomes increasingly dangerous with the number of repetition of the game. If ambushed, losses are much more important with this kind of planners. Overall these evaluations show that the method described in this paper in more adapted to solve ambush games. Some enhancement can be included through the definition of secondary objectives.

\section{Future Work}

For the purpose of this paper the risk assessment map, i.e. assigning value for \textit{alpha} at each node, was constructed arbitrarily. In order to get this technique one step closer to a practical application, research is being pursued on strategic field analysis for realistic risk assessment. The first step, currently in development, is to be able to analyse real 3D GPS data and understand how the topological configuration of the environment will influence the amount of damages inflicted to the convoy if ambushed. Once able to efficiently forecast the casualties from their correlation with the environment, this method will represent a good high-level planner for vehicles moving in hostile environments. The second step is to develop a lower level planner based on real-time risk assessment that might come from video flow analysis and threat detection. This threat could then be by-passed using high speed manoeuvres control algorithm currently investigated at Georgia Tech and MIT under the MURI \textit{"Neuro-Inspired Event-Driven Perception and Control of Autonomous 
Vehicles for Aggressive Driving"}.

Among other subjects in need for further investigation is the importance of the degree of information Player 1 and Player 2 share. It was assumed in Section IV that Player 2 had perfect knowledge of the strategy of Player 1 for this method, or that he could guess it for the two deterministic planners. This idea of "revealing" and "non-revealing" strategies is currently under investigation by Jones \& Shamma (Georgia Tech) and would represent a good addition to the game theoretic aspect of this method.

Finally, more attention will be brought to the specificities of the environment discretization. The node-ambush relation, for example, need to be developed. Currently, if Player 2 is at node (1,1) and Player 1 goes through node (1,2) then Player 1 cannot be ambushed. But in an unstructured environment it might be possible that Player 2 intercepts Player 1 if the nodes are close enough. Keeping in mind practical applications, this type of situations has to be studied more deeply.

\section{Conclusion}
A method for probabilistic path planning using game theory and Linear Programming was presented. Several methods of discretization of the environment are presented and compared. While both Method 2 and Method 3 (cf. Section III.A) result in a good representation of the environment, the later one is chosen for its computational power efficiency.

Though this method had been studied before, an important feature was presented, which is the sensitivity of the type of solution returned by the method to the LP solver used. The entropy of the resulting probability distribution is explained to be a good indicator of the quality of the solution for the purpose of ambush games and deception. The Interior Point algorithm is preferred over the Simplex algorithm because of the higher entropy of its solutions.

The energy efficiency of this method being relatively low, a secondary objective is proposed for the linear optimization that allows some trade-off between safety and energy efficiency.
In the end, this method is shown to offer more safety than traditional deterministic planners, especially for repetitive journeys. This feature is the most important, as it is meant to increase the security of convoys having such a dangerous routine.

\section*{Acknowledgment}
This work was supported by the Army Research Office under MURI Award
W911NF-11-1-0046.



%

\end{document}